\pdfoutput=1

\documentclass[11pt]{article}

\usepackage{ACL2023}

\usepackage{times}
\usepackage{placeins}

\usepackage{latexsym}
\usepackage{multirow}
\definecolor{mynavy}{RGB}{0, 0, 128}
\definecolor{mypurple}{RGB}{128, 0, 128}
\definecolor{myturquoise}{RGB}{64, 224, 208}
\usepackage{amsmath}
\usepackage{algorithm}
\usepackage{algpseudocode}

\usepackage{bbding}
\usepackage{pifont}
\usepackage{wasysym}
\usepackage{amssymb}
\usepackage{booktabs}
\usepackage[multiple]{footmisc}
\usepackage{xcolor}
\usepackage[export]{adjustbox}

\usepackage{listings}

\usepackage[colorinlistoftodos]{todonotes}
\usepackage{subcaption}
\usepackage{url}
\usepackage{array}
\usepackage[utf8]{inputenc}
\usepackage{float}
\usepackage{multicol}

\usepackage[T1]{fontenc}
\usepackage[utf8]{inputenc}
\usepackage{microtype}
\usepackage{inconsolata}

\title{GECTurk WEB: An Explainable Online Platform for Turkish Grammatical Error Detection and Correction}

\author{Ali Gebeşçe\textsuperscript{$1, 2$},~
Gözde Gül Şahin\textsuperscript{$1, 2$}
\\[.3em]
\textsuperscript{1}Computer Engineering Department, Koç University, Istanbul, Turkey\\
\textsuperscript{2} KUIS AI Lab, Istanbul, Turkey \\
{\url{https://gglab-ku.github.io/}}\\
}

\begin{document}
\maketitle
        
\begin{abstract}
    Sophisticated grammatical error detection/correction tools are available for a small set of languages such as English and Chinese. However, it is not straightforward---if not impossible---to adapt them to morphologically rich languages with complex writing rules like Turkish which has more than 80 million speakers. Even though several tools exist for Turkish, they primarily focus on spelling errors rather than grammatical errors and lack features such as web interfaces, error explanations and feedback mechanisms. To fill this gap, we introduce \textsc{GECTurk WEB}, a light, open-source, and flexible web-based system that can detect and correct the most common forms of Turkish writing errors, such as the misuse of diacritics, compound and foreign words, pronouns, light verbs along with spelling mistakes. Our system provides native speakers and second language learners an easily accessible tool to detect/correct such mistakes and \textit{also to learn from their mistakes} by showing the explanation for the violated rule(s). The proposed system achieves 88,3 system usability score, and is shown to help learn/remember a grammatical rule (confirmed by 80\% of the participants). The \textsc{GECTurk WEB} is available both as an offline tool~\footnote{\tiny{\url{https://github.com/GGLAB-KU/gecturkweb}}} or at \url{www.gecturk.net}. 
\end{abstract}

\section{Introduction}
    


    Grammatical Error Correction/Detection (GEC/D)~\cite{bryant-survey} is a well-established NLP task, that aims to detect and correct various errors in text, including grammatical issues like missing prepositions, mismatched subject-verb agreement, as well as orthographic and semantic errors such as misspellings and inappropriate word choices. Tools that can perform GEC/D have recently gained attention due to the rise in digital communication, remote work, and global interactions, which demand clear and professional writing. With the inclusion of the detection module, GEC/D formulation facilitates the \textit{teaching of grammar rules}, empowering users not only to produce error-free writing but also to enhance their language skills and comprehension gradually.

    Therefore, developing open-source GEC/D tools is particularly crucial, yet challenging for languages with complex writing rules, such as Turkish. The writing rules for such languages generally involve multiple linguistic layers---phonetic, syntactic, and semantic---which makes them difficult to follow and remember even for native speakers. While several tools exist for high-resource languages such as GECko+~\cite{gecko+} and ALLECS~\cite{en-5}, they often suffer from discontinuation of support or lack adaptability for languages such as Turkish. Moreover, while advanced commercial tools such as LanguageTool\footnote{\tiny \url{https://languagetool.org/}} offer support for 31 languages, yet Turkish is notably absent from their list. Furthermore, as highlighted in \S~\ref{sec:prev}, numerous offline tools are accessible for Turkish spelling correction, whereas only two models (not tools)~\cite{uz-eryigit-2023-towards,kara-etal-2023-gecturk} are dedicated to Turkish GEC/D.

    \begin{figure*}
    \centering
    \includegraphics[width=\textwidth]{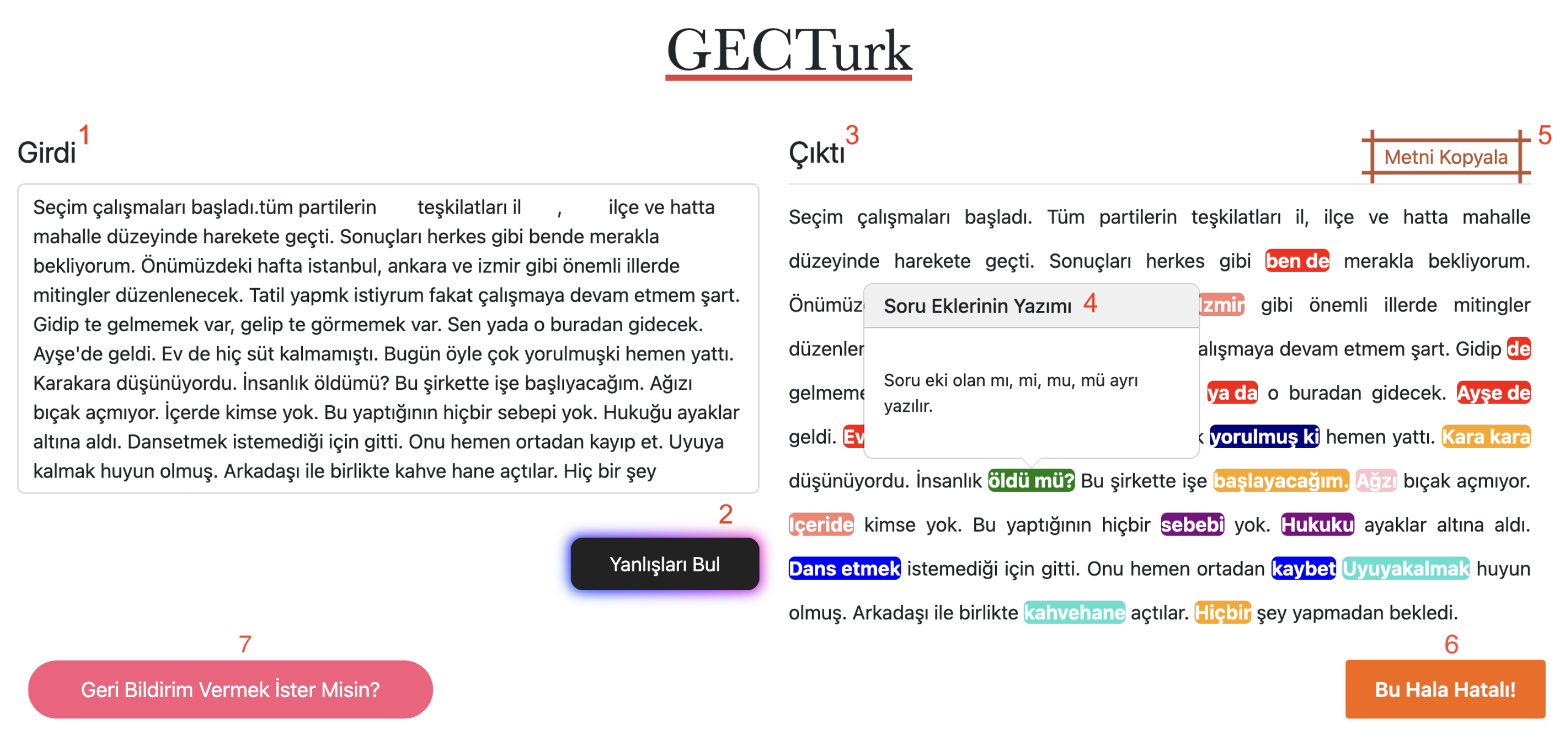}
        \caption{The screenshot of UI after user entering an input. \textbf{1-} Girdi (Input): The input area for the user. \textbf{2-} Yanlışları Bul (Find Errors): A button which is pressed after entering an input. \textbf{3-} Çıktı (Output): The output area for the tagged and corrected text. Note that each error is categorized (colored) according to Table~\ref{table:2}. \textbf{4-} Pop-up: Each corrected word is represented as button. When clicked the violated rule, i.e., error type, is shown. \textbf{5-} Metni Kopyala (Copy Text): A button for copying corrected text. \textbf{6-} Bu Hala Hatalı (Still Erroneous): A button for giving feedback in case the user thinks the output still contains errors. When clicked, a pop-up is shown and user is expected to write the corrected version. \textbf{7-} Geri Bildirim Vermek İster Misin? (Give Feedback): A button for collecting general suggestions.}
    \label{fig:before-after}
    \end{figure*}
    
    
    
    To bridge this gap, we leverage the state-of-the-art pretrained GEC/D~\cite{kara-etal-2023-gecturk}\footnote{We use the pretrained sequence tagging model that has been trained on 130,000 high-quality sentences covering more than 20 expert-curated grammar rules (a.k.a., writing rules) implemented through complex transformation functions.} and spelling correction models; and, for the first time, provide a user-friendly web-interface to them. Our system does not only correct errors but also display them in different colors, while providing explanations for each correction through interactive elements in the interface. Additionally, the system includes a feedback mechanism to foster continuous improvement and enhance user engagement. Our system is lightweight and flexible, allowing easy adaptation to other languages through pretrained sequence tagging models. The results of the user study (see \S\ref{sec:evaluation}) demonstrate excellent usability and a significant impact on learning and retention of grammar rules. \textsc{GECTurk WEB}, shown in Fig.~\ref{fig:before-after}, is accessible both as an offline tool and online at \url{www.gecturk.net} and source code licensed with CC BY-SA 4.0 is available at \url{https://github.com/GGLAB-KU/gecturkweb}. 
    
    

	\begin{table*}[h!]
		\centering
        \scalebox{0.6}{
	    \begin{tabular}{cccccccc} 
	        \toprule
	        & Spelling & Offline & Open Source & Grammatical & Explanation & Feedback & Web Interface \\
	        \midrule
            Google Docs & \checkmark &  &  &  & & & \checkmark \\
         	Microsoft Word & \checkmark & \checkmark &  &  & & & \checkmark \\
         	Zemberek~\cite{zemberek} & \checkmark & \checkmark & \checkmark & & & & \\
            Hunspell~\cite{hunspell} & \checkmark & \checkmark & \checkmark & & & & \\
         	TurkishNLP\cite{turkishnlp} & \checkmark & \checkmark & \checkmark & & & & \\
         	TrNLP~\cite{trnlp} & \checkmark & \checkmark & \checkmark & & & & \\
            Starlang~\citet{starlang} & \checkmark & \checkmark & \checkmark & & & & \\
            VNLP~\cite{vnlp} & \checkmark & \checkmark & \checkmark & & & & \checkmark \\
         	Mukayese~\cite{safaya2022mukayese} & \checkmark & \checkmark & \checkmark & & & & \\
            \midrule
	        Rule-based~\cite{uz-eryigit-2023-towards} & \checkmark & \checkmark & \checkmark & \checkmark & \checkmark & & \\
            GECTurk~\cite{kara-etal-2023-gecturk} &  & \checkmark & \checkmark & \checkmark & \checkmark & & \\
	        GECTurk WEB (Ours) & \checkmark & \checkmark & \checkmark & \checkmark & \checkmark & \checkmark& \checkmark\\
	        \bottomrule
	        \end{tabular}
         }
            \caption{Comparison of features in previous grammatical and spelling error correction tools for Turkish, contrasted with ours. Spelling: Correction of spelling errors. Grammatical: Detection of grammatical errors. Explanation: Explanations for error types. Feedback: User feedback mechanism for model and interface improvement. Web Interface: Availability of a web-based interface.}
	    \label{table:1}
	\end{table*}

\section{Previous Systems} 
\label{sec:prev}
    \paragraph{High-Resource Languages:} 
    Numerous GEC/D models exist for high-resource languages such as English \cite{en-1, en-2, en-3, en-4} and Chinese \cite{ch-1, ch-2, ch-3, ch-4}. However, these models lack user interfaces, which are crucial for accessibility to non-specialists. Although fewer in number compared to models, several GEC/D tools are available. For instance, GECko+ \cite{gecko+} integrates the GECToR XLNet model for sentence-level grammatical correction with a sentence ordering model \cite{sent-order}. It processes texts by segmenting them into sentences, applying corrections, and then reordering them. Initially, GECko+ offered a web interface, but it is currently inactive. Now, the only access is through downloading the source code and running it locally, which is inconvenient for general users. Similarly, MiSS \cite{miss}, a Multi-Style Simultaneous Translation system that includes a GEC/D feature using GECToR XLNet, initially had a web interface which is now inactive.
    
    The most recent non-commercial GEC/D tool is ALLECS \cite{en-5}, which uses GECToR-RoBERTa, GECToR-XLNet, and T5-Large models, alongside two combination methods: ESC \cite{esc} and MEMT \cite{memt}. ALLECS takes input and displays corrected errors with clickable buttons, and has an easy-to-use web interface. Despite its advantages, ALLECS lacks a feedback mechanism and an enhanced interface that uses color coding to distinguish between different types of errors. Moreover, its implementation is not flexible enough to be extended to other languages, i.e., one cannot simply upload a Turkish GEC/D model and expect the application to function without significant modifications to the source code.

    \paragraph{Morphologically Rich Languages:}
    In the case of morphologically rich languages, there are fewer GEC/D models available. Examples include Arabic \cite{ar-1}, Bengali \cite{ben-1}, Czech \cite{cz-1, cz-2}, and Russian \cite{ru-1}. However, again these systems lack user interfaces, making them merely as models rather than practical tools, thus limiting their usability for general users. One exception exists in Arabic; however, this tool just underlines mistakes \cite{ar-2} and not explain the errors. Also it lacks a web support, making it less suitable for general users.
    
    \paragraph{Commercial Tools:} 
    Grammarly\footnote{\tiny \url{https://www.grammarly.com/}} offers advanced features for improving writing tone on several aspects like clarity, engagement, and delivery. However, it is not open-source and supports only English. Also, full access to its features requires a paid subscription \footnote{\tiny \url{https://www.grammarly.com/plans}}. LanguageTool, being open-source, supports multiple languages and addresses some of Grammarly's limitations. However, it imposes a 10,000-character limit on inputs, expandable only through a paid subscription \footnote{\tiny \url{https://languagetool.org/premium_new}}. More importantly, despite supporting 31 languages \footnote{\tiny \url{https://dev.languagetool.org/languages}}, Turkish is not among them.
    
    \paragraph{Turkish:} Since aforementioned systems are either commercial or not directly applicable to Turkish GEC/D, we have surveyed commonly available tools and resources that offer support for Turkish, given in Table~\ref{table:1}. Google Docs \footnote{\tiny\url{https://docs.google.com}} and Microsoft Word \footnote{\tiny\url{https://www.microsoft.com/word}}, widely accessible for their user-friendly interfaces, provide basic spelling error detection. However, they fall short in addressing the specific grammatical nuances of the Turkish language. Additionally, these tools are not open-source, lack explanations for corrections, and do not offer a mechanism for user feedback. There are also open-source tools, such as Zemberek~\cite{zemberek}, Hunspell~\cite{hunspell}, TurkishNLP~\cite{turkishnlp}, TrNLP~\cite{trnlp}, StarlangSoftware~\cite{starlang}, VNLP~\cite{vnlp} and MukayeseSpellChecker~\cite{safaya2022mukayese}, however they only provide an offline spelling. To the best of our knowledge, there are only two resources for Turkish GEC/D~\cite{uz-eryigit-2023-towards, kara-etal-2023-gecturk}. \citet{uz-eryigit-2023-towards} propose a rule-based, offline GED system that employs common, universal error types \cite{bryant-etal-2017-automatic}, while \citet{kara-etal-2023-gecturk} provide several pre-trained GEC and GED models that can detect expert-curated language specific writing rules and show significant improvements on existing and proposed benchmarks. In this work, we combine the state-of-the-art GEC/D~\cite{kara-etal-2023-gecturk} and spelling correction models; and, for the first time, provide a user-friendly web-interface to them. Additionally, we provide colorful explanations for a wide range of error types to train the users, and incorporate a feedback mechanism for continuous training of pre-trained models.    

\section{\textsc{GECTurk WEB}}
\label{sec:model}
    Our system has four main components: i) frontend, ii) backend, iii) grammatical error correction/detection (GEC/D), and iv) spelling correction modules. \textsc{GECTurk WEB} is based on the Python Django framework,\footnote{\tiny \url{https://www.djangoproject.com}} which manages everything related to performance, security, scalability, and database handling. The architecture of our system, incorporating these components along with the data flow, is shown in Figure~\ref{fig:fig2}. 
    
    \begin{table}
    \centering
    \scalebox{0.5}{
        \begin{tabular}{
          >{\centering}m{1cm}
          >{\centering\arraybackslash}m{3cm}
          >{\centering\arraybackslash}m{4cm}
          >{\centering\arraybackslash}m{3cm}
          >{\centering\arraybackslash}m{2cm}
        }
        \toprule
        \textbf{Category} & \textbf{Rule ID} & \textbf{Description} & \textbf{Example Correction} & \textbf{Color} \\
        \toprule
        \rotatebox{90}{-DE/-DA} & 1. CONJ\_DE\_SEP & Conjunction “-de/-da” is written separately. & Durumu [oğlunada $\rightarrow$ oğluna da] bildirdi. & \textcolor{red}{\textbf{Red}} \\
        \midrule
        \rotatebox{90}{-KI} & 7. CONJ\_KI\_SEP & Conjunction “-ki” is written separately. & Bugün öyle çok [yorulmuşki $\rightarrow$ yorulmuş ki] hemen yattı. & \textcolor{mynavy}{\textbf{Navy}} \\
        \midrule
        \rotatebox{90}{FOREIGN} & 9. FOREIGN\_R1 & Words that start with double consonants of foreign origin are written without adding an “-i” between the letters. & [gıram $\rightarrow$ gram] & \textcolor{mypurple}{\textbf{Purple}} \\
        \midrule
        \rotatebox{90}{BISYL} & 13. BISYLL\_HAPL\_VOW & Some bisyllabic words undergo haplology when they get a suffix starting with a vowel. & [ağızı $\rightarrow$ ağzı]& \textcolor{pink}{\textbf{Pink}} \\
        \midrule
        \rotatebox{90}{LIGHT VERB} & 17. LIGHT\_VERB\_SEP & Light verbs such as “etmek, edilmek, eylemek, olmak, olunmak” are written separately in case of no phonological assimilation &  [arzetmek $\rightarrow$ arz etmek] & \textcolor{blue}{\textbf{Blue}} \\
        \midrule
        \rotatebox{90}{COMPOUND} & 20. COMP\_VERB\_ADJ & Compound words formed by knowing, giving, staying, stopping, coming, and writing are written adjacent if they have a suffix starting with -a, -e, -ı, -i, -u, -ü. & [uyuya kalmak $\rightarrow$ uyuyakalma], [gide durmak $\rightarrow$ gidedurmak] & \textcolor{myturquoise}{\textbf{Turquoise}} \\
        \midrule
        \rotatebox{90}{SINGLE} & 22. PRONOUN\_EXC & Traditionally, some pronouns are written adjacent. & [hiç bir $\rightarrow$ hiçbir], [her hangi $\rightarrow$ herhangi] & \textcolor{orange}{\textbf{Orange}} \\
        \bottomrule
        \end{tabular}
    }
    \caption{A selection of grammatical error types covered in the system from \citet{kara-etal-2023-gecturk}.}
    \label{table:2}
    \end{table}

    \subsection{Frontend}
    \label{ssec:frontend}
        For the user interface, we use the Bootstrap framework~\footnote{\tiny \url{https://getbootstrap.com}} that provides us with modern, responsive, and mobile compatible HTML and CSS. Initially, empty ``Input'' and ``Output'' fields are shown. After identifying and correcting grammatical and spelling errors in the input, the output is enriched with error types (see Figure~\ref{fig:before-after}). For each correction, HTML snippets are created to wrap the corrected words and transforms them to actionable buttons. These snippets use Bootstrap's pop-over functionality to provide an interactive way to display the error type, an explanation, and the correction. Each correction is highlighted with a specified background color and font size for visibility. Additional information about each error type is retrieved from a predefined set of rules given in Table~\ref{table:2}\footnote{We refer the readers to \citet{kara-etal-2023-gecturk} for details on each writing rule and how they are handled by the model.}. This information includes a textual explanation and a title for the error, which are both used in the content of the pop-over. For instance, if there is a misspelling of ``-de/da'', this is displayed as \textit{Conjunction ``-de/da'' is always written separately}. The tokens within the input text are replaced with the generated HTML snippets, respecting the original positions of errors. This involves calculating the offsets to accurately place the HTML snippets within the text, considering the length of the corrected phrases. The corrected tokens are joined back together into strings for each line, and then all lines are combined into a single HTML paragraph (<p> tags). 

    \subsection{Backend}
    \label{ssec:backend}
        Our system uses Django, a high-level Python web framework, to create a strong backend infrastructure. The architecture of Django, known as Model-View-Template (MVT), supports a clear separation of responsibilities. Here, the Model is responsible for data storage and retrieval. The View handles user requests and provides responses, and the Template dynamically generates HTML pages for user interaction.
        
        \paragraph{View} The \texttt{send\_data} function is used for accommodating various actions including text submission for correction, feedback submission, and API interactions. Upon receiving a POST request given the input text, the function invokes a text correction process through \texttt{get\_text\_corrector}. Text correction process starts with sentence tokenization using NLTK's \texttt{sent\_tokenize} function \cite{bird2009natural} and continues with the grammatical error correction process, which is described in detail in \S\ref{ssec:gc}. The corrected text, alongside original input and HTML-formatted output for interactive display, is then encapsulated within a \textsc{Text} model instance for persistence. Feedback submission, whether specific to text corrections or general website feedback, is similarly processed and stored.

        \paragraph{Model} Our data model chas two main entities: \textsc{Text} and \textsc{GeneralFeedback}. The \textsc{Text} model captures the essence of each correction session, storing original and corrected texts, HTML-tagged corrected text for frontend display, and any user feedback. This allows for a comprehensive audit trail of user interactions and system outputs. The \textsc{GeneralFeedback} model, on the other hand, aggregates general user impressions and feedback about the website, enabling continuous improvement based on user insights.
        
        \paragraph{Database and Server} Thanks to Django's ORM capabilities, we easily integrate these models with our MySQL\footnote{\tiny \url{https://www.mysql.com/}} database, as the database management system. 
        We use AWS Elastic Beanstalk\footnote{\tiny \url{https://docs.aws.amazon.com/elasticbeanstalk/latest/dg}} for deployment.

        \begin{figure*}
        \centering
        \includegraphics[width=\textwidth]{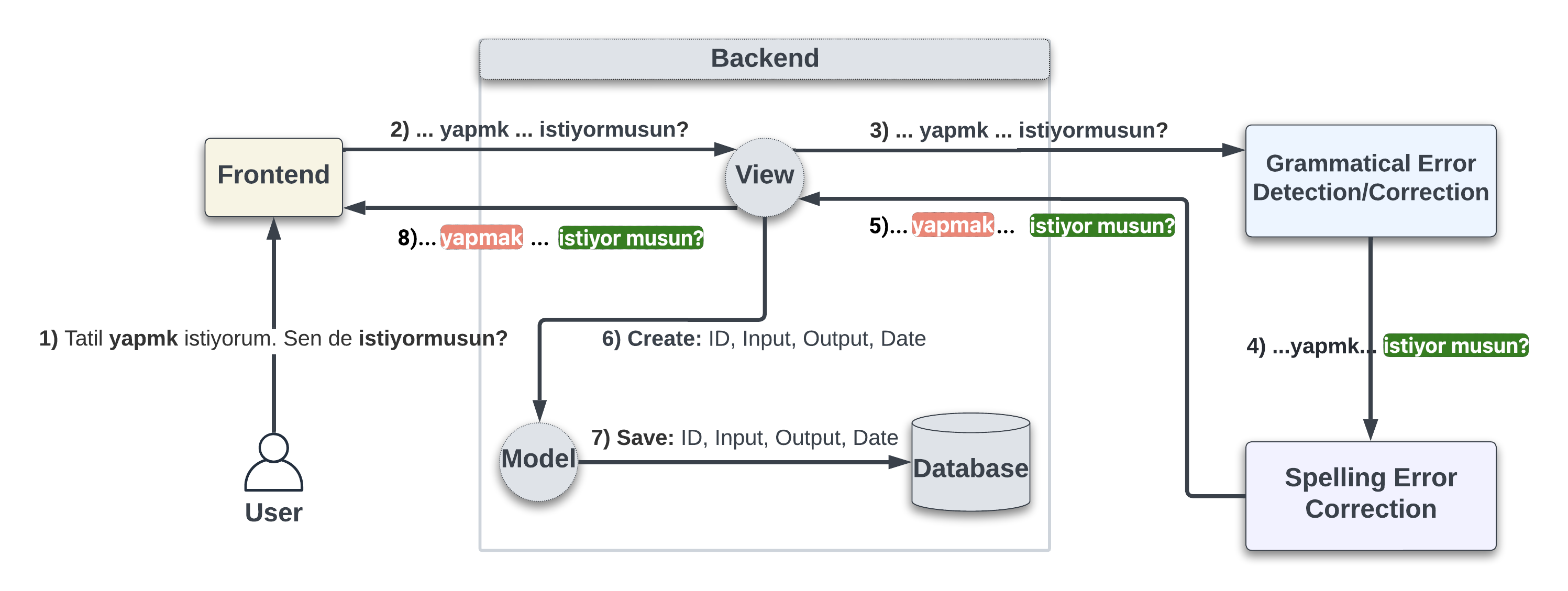}
        \caption{The \textsc{GECTurk WEB} Architecture. \textbf{1)} User inputs text containing two errors: a spelling error, ``\textbf{yapmk}'' (shown in red) and a grammatical error, ``\textbf{istiyormusun}'' (shown in green). \textbf{2-3)} The view receives the input from the frontend and forwards it to the GEC/D model. \textbf{4)} The GEC/D model corrects the grammatical error and adds tags for the frontend to display, as shown in \ref{fig:before-after}. \textbf{5)} The SEC module corrects the spelling error, tags it, and sends it back to the View. \textbf{6-7)} The model compiles relevant information such as ID, Input, Output, and Date, and records these in the database. \textbf{8)} The View sends the prepared output back to the frontend for display.}
        \label{fig:fig2}
    \end{figure*}

    \subsection{Grammatical Correction}
    \label{ssec:gc}
        We employ the state-of-the-art GEC/D model, SequenceTagger, previously described in \citet{kara-etal-2023-gecturk}. 
        Briefly, SequenceTagger finetunes a strong encoder model (e.g., BERTurk~\cite{stefan_schweter_2020_3770924}) to classify tokens into grammatical error classes, enabling efficient error detection rather than merely correction. For illustrative purposes, we provide one sample error type from each category in Table~\ref{table:2}. Then, corrections are performed with reverse transformations. 
        The model weights and associated files, such as the tokenizer and vocabulary, are securely stored on Amazon S3\footnote{\tiny \url{https://aws.amazon.com/s3}}. Deployment is simplified through the use of AWS Elastic Beanstalk, requiring only the compression of the project (including the model itself) and uploading it to the AWS Elastic Beanstalk application. We have adapted the original code from \cite{kara-etal-2023-gecturk} into a class named \textsc{TextCorrector} and an API function \texttt{process\_text} for performing correction operations with this model. For further details, we encourage consulting the source code of \citeauthor{kara-etal-2023-gecturk} \footnote{\tiny \url{https://github.com/GGLAB-KU/gecturk}} and our implementation \footnote{\tiny \url{https://github.com/GGLAB-KU/gecturkweb}}.
                            
    \subsection{Spelling Correction}
    \label{ssec:sc}

        It should be noted that users not only make grammatical mistakes but also commonly commit spelling errors. Since the GEC/D model is not designed for spelling error correction, we employ external tools to extend our system. For mistakes in proper nouns and common typos, we survey external Turkish spelling correction tools. After evaluating different options, we find VNLP~\cite{vnlp}, StarlangSoftware~\cite{starlang}, and TurkishNLP~\cite{turkishnlp} unsatisfactory by means of efficiency and accuracy. As a result, we integrated TrNlp~\cite{trnlp} and ZemberekNLP \cite{akin2007zemberek, zemberek, zemberekpy} to our system.  We apply corrections using TrNlp for proper noun capitalization (e.g., ``ankara'' $\rightarrow$ ``Ankara'')---e.g., any proper noun violating it is capitalized by the tool. Following the proper noun corrections, we leverage ZemberekNLP's \textsc{TurkishSentenceNormalizer} for the common typos. Sentences are processed to ensure that the words are not corrupted (e.g., ``yapmk'' $\rightarrow$ ``yapmak'') and that consistency is maintained across the text. With the combination of TrNlp and ZemberekNLP, our system now not only fixes grammatical errors but also performs spelling correction in Turkish.
        
    \section{Evaluation}
    \label{sec:evaluation}
    
    To evaluate \textsc{GECTurk WEB}, we conduct an in-depth user study. This study aims to assess the usability and effectiveness of the tool in facilitating learning and retention.

    The user study is structured into two parts. First, participants are asked to follow a user scenario, where they input 10 short sentences into \textsc{GECTurk WEB}. These sentences are selected to cover all four possible outcomes: True Positives (TP), where the system accurately identifies and corrects an error; True Negatives (TN), where no error exists and the system appropriately refrains from making changes; False Positives (FP), where the system erroneously alters a correct sentence; and False Negatives (FN), where the system overlooks an error. Reflecting on the performance of \textsc{GECTurk}~\cite{kara-etal-2023-gecturk}, which demonstrated a detection precision of 0.89 and a correction F1-score of 0.84, we have designed a representative sample to mirror these results. Therefore, the set of 10 sentences includes 7 True Positives (TPs) and 1 of each other outcome types. It is important to note that the participants are unaware of this distribution. To guide the participants on each potential outcome, we create four videos and present them to participants before they begin experimenting with \textsc{GECTurk WEB}, which is described in detail in \S\ref{ssec:first-part}. After viewing these videos, participants are instructed to input each sentence and classify it according to one of the possible outcomes. The complete list of 10 sentences can be found in \S\ref{ssec:first-part}. We restrict the average duration of this part to be 45 minutes to align with findings from studies \cite{lavrakas2008encyclopedia, Kost2018ImpactOS, sharma2022short} on the optimal length for questionnaires.
    After completing this part, participants are asked several questions to assess the system based on the evaluation metrics. We employ two established metrics to test usability and user satisfaction: the System Usability Scale (SUS)~\cite{sus} and the Standardized User Experience Percentile Rank Questionnaire (SUPR-Q)~\cite{suprq}. These metrics are widely recognized for their reliability in assessing user satisfaction and system usability. To understand the effectiveness of \textsc{GECTurk WEB}, we also ask a yes/no question about whether participants learned or remembered a grammatical rule. The SUS questionnaire contains ten five-level Likert scale questions. The SUPR-Q includes seven five-level and one ten-level Likert scale questions. Including our yes/no question, we ask a total of 19 questions. All of these questions are in \S\ref{ssec:second-part}.
    
    The evaluation results from 10 users are noteworthy, particularly in terms of usability and user satisfaction. The average SUS score is 88.3 (out of a possible 100; the average benchmark is 69 \cite{susaverage}), indicating an excellent level of usability. Similarly, the average score for the SUPR-Q was 4.34 (out of a possible 5; the average benchmark is 3.93 \cite{suprq}), suggesting high user satisfaction with the web interface and functionality. These scores are significantly above the average benchmarks, highlighting the effectiveness of \textsc{GECTurk WEB} in providing a user-friendly and satisfying experience. Notably, 80\% of participants report that they learned or remembered a grammatical rule, underscoring the tool's impact on learning and retention. 
    Additionally, we measure the time-efficiency of the system and provide the results in Appendix \S\ref{sec:time-eff-app}.
    
\section{Extension to Other Languages}
\label{sec:ext}

As depicted in Figure \ref{fig:fig2}, our system exhibits flexibility and seamless adaptability for multilingual support. Expanding our system to support other languages merely requires the replacement of the GED/C model and the spelling error correction module. Specifically, the sequence tagger model must be trained to identify the distinct grammatical error patterns of the target language. Similarly, the spelling error correction module can be replaced with an existing spelling corrector for the target language. Both modules can be adjusted by modifying the ``text\_corrector.py'' script and the associated model weights files, facilitating straightforward integration.

\section{Conclusion}
\label{sec:conc}

    In this work, we present \textsc{GECTurk WEB}, a practical online platform for Turkish grammatical error detection and correction (GED/C) along with spelling error correction (SEC). Our system aims to not only correct mistakes but also to facilitate learning of complex writing rules via user-friendly rule explanations. Furthermore, the user feedback mechanism allows for continual support and training of the tool. 
    The high SUS and SUPR-Q scores, significantly above average benchmarks, alongside the positive feedback on learning outcomes, validate the platform's design philosophy and its focus on user-centric development. Furthermore, \textsc{GECTurk WEB} is built with a flexible architecture, suggesting that adaptation to additional languages is within reach. Source code and the web-based tool is publicly and freely available. 

\section*{Limitations}
    Major limitation of our system is the number of concurrent user interactions it can process. Currently, the system operates on a single AWS i4i.large instance, which can efficiently manage up to ten simultaneous users. Beyond this threshold, performance begins to degrade, necessitating additional instances to preserve service quality. However, it's essential to highlight that this limitation can easily be overcome by enhancing our infrastructure given the budget. Should the \textsc{GECTurk WEB}  platform experience a surge in popularity, we are prepared to scale our resources horizontally by incorporating more instances.

\section*{Ethics Statement}
    The development and deployment of \textsc{GECTurk WEB} adhere to ethical considerations crucial for language processing tools. We ensure that user data is handled with the utmost confidentiality and integrity, in accordance with data protection regulations. The feedback system is designed to be non-intrusive and respectful of user privacy.

\section*{Acknowledgements}
We thank Asu Tutku Gökçek, Gökçe Sevimli, and Yakup Enes Güven for their cnotributions to the project. This work has been supported by the Scientific and Technological Research Council of Türkiye~(TÜBİTAK) as part of the project ``Automatic Learning of Procedural Language from Natural Language Instructions for Intelligent Assistance'' with the number 121C132. The authors also gratefully acknowledge KUIS AI Lab for providing computational support.

\bibliography{custom}
\bibliographystyle{acl_natbib}

\appendix
\clearpage
\section*{Appendix} 
\label{sec:appendix}
\section{User Study}

\subsection{The user scenario}
\label{ssec:first-part}

\begin{figure}[!htb]
\centering
\fbox{\includegraphics[width=0.39\textwidth]{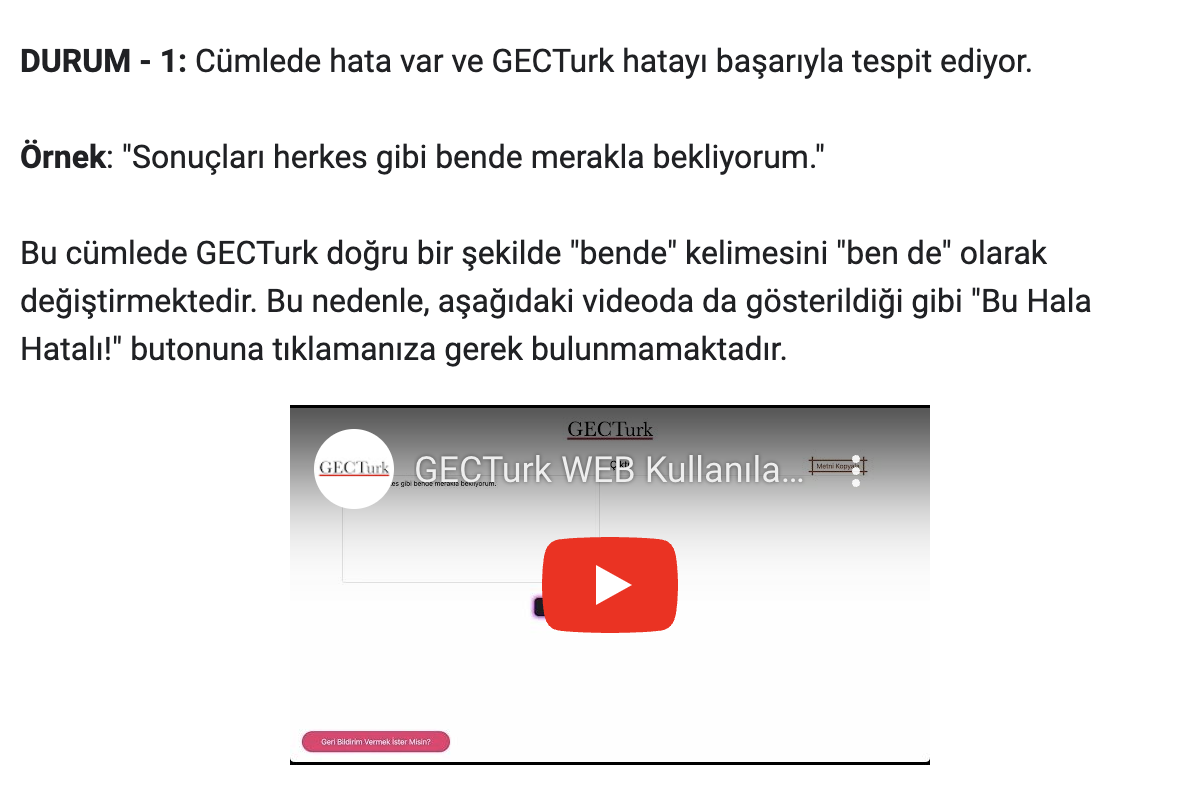}}
\caption{\textbf{CASE - 1:} The sentence contains an error and GECTurk successfully detects the error. \textbf{Example:} ``Sonuçları herkes gibi bende merakla bekliyorum.'' In this sentence, GECTurk correctly changes ``bende'' to ``ben de''. Therefore, there is no need to click on the ``This is still incorrect!'' button, as shown in the video below.}
\label{fig:case-1}
\end{figure}

\begin{figure}[!htb]
\centering
\fbox{\includegraphics[width=0.39\textwidth]{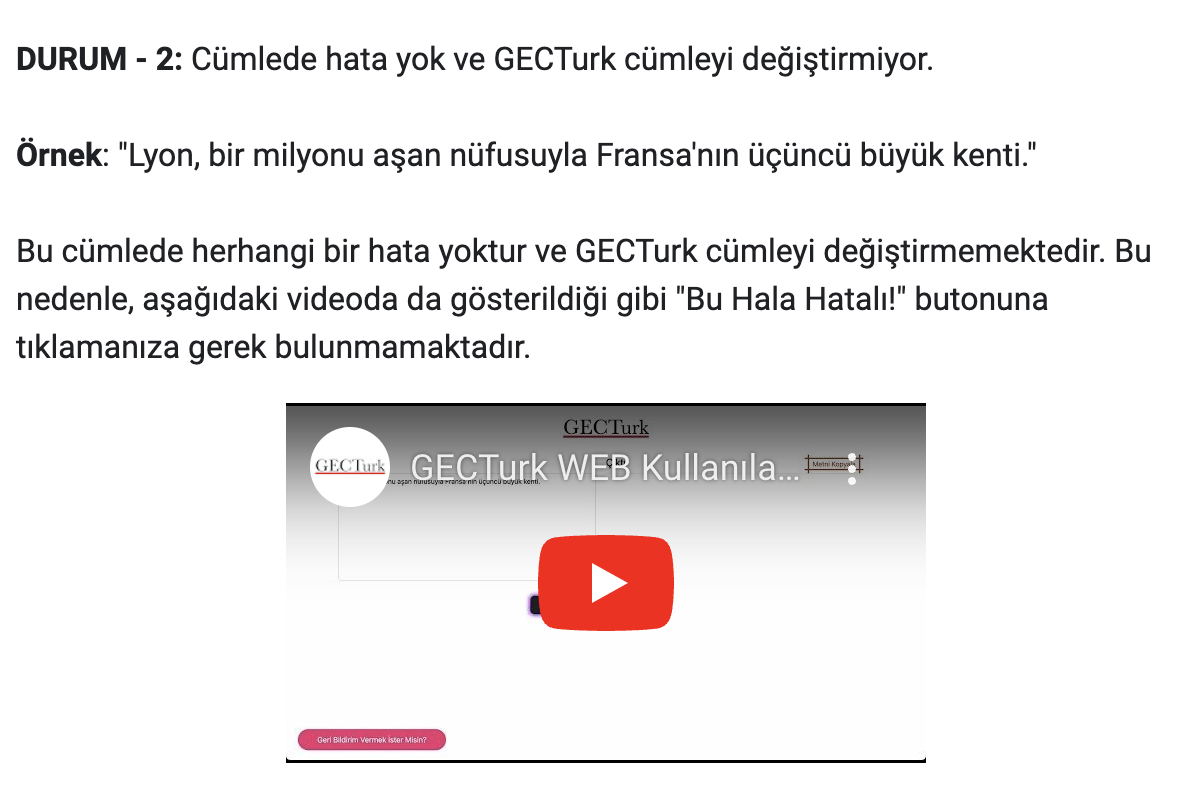}}
\caption{\textbf{CASE - 2:} There is no error in the sentence and GECTurk does not change the sentence. \textbf{Example:} ``Lyon, bir milyonu aşan nüfusuyla Fransa'nın üçüncü büyük kenti.'' There are no errors in this sentence and GECTurk does not change the sentence. Therefore, there is no need to click on the ``This is still incorrect!'' button, as shown in the video below. }
\label{fig:case-2}
\end{figure}

\begin{figure}[!htb]
\centering
\fbox{\includegraphics[width=0.39\textwidth]{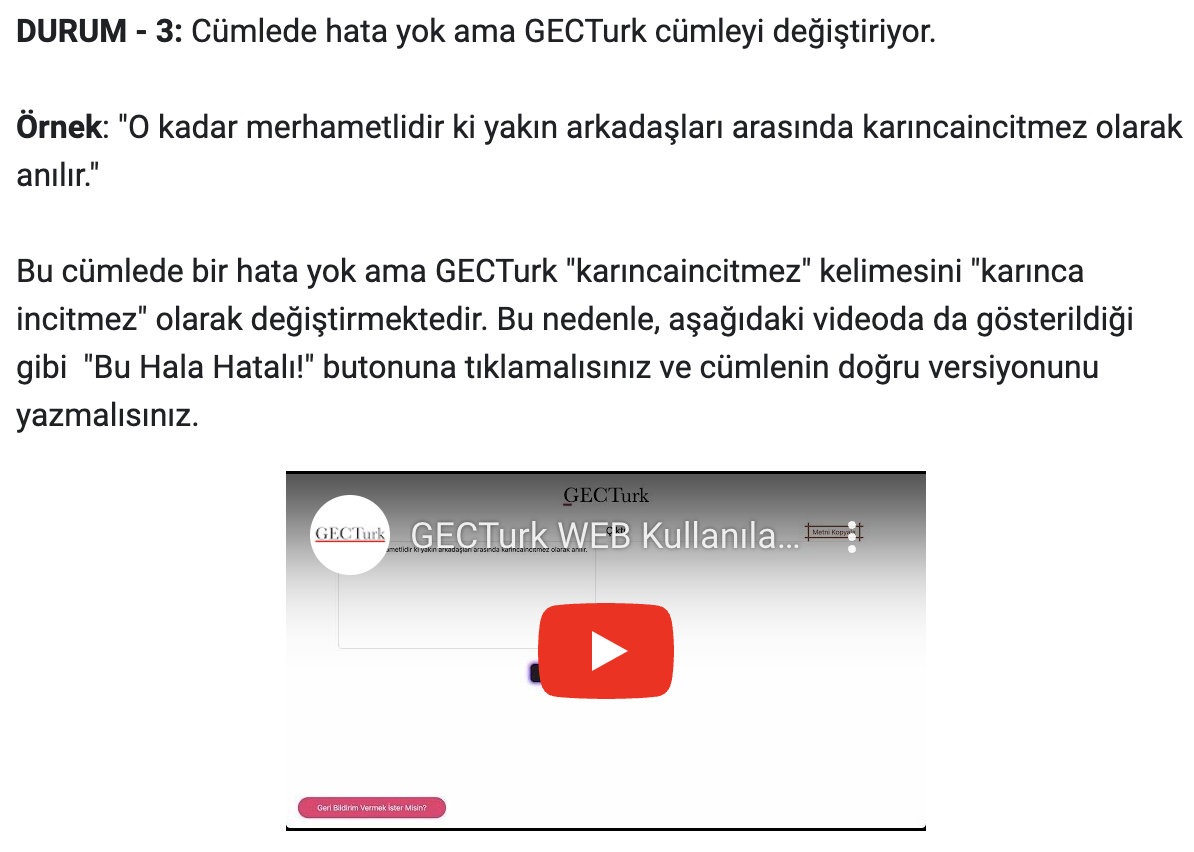}}
\caption{\textbf{CASE - 3:} There is no error in the sentence but GECTurk changes the sentence. \textbf{Example:} ``O kadar merhametlidir ki yakın arkadaşları arasında karıncaincitmez olarak anılır.'' There is no mistake in this sentence, but GECTurk changes the word ``karıncaincitmez'' to ``karınca incitmez''. Therefore, you should click on the ``This is still incorrect!'' button and type the correct version of the sentence, as shown in the video below.}
\label{fig:case-3}
\end{figure}

\begin{figure}[!htb]
\centering
\fbox{\includegraphics[width=0.39\textwidth]{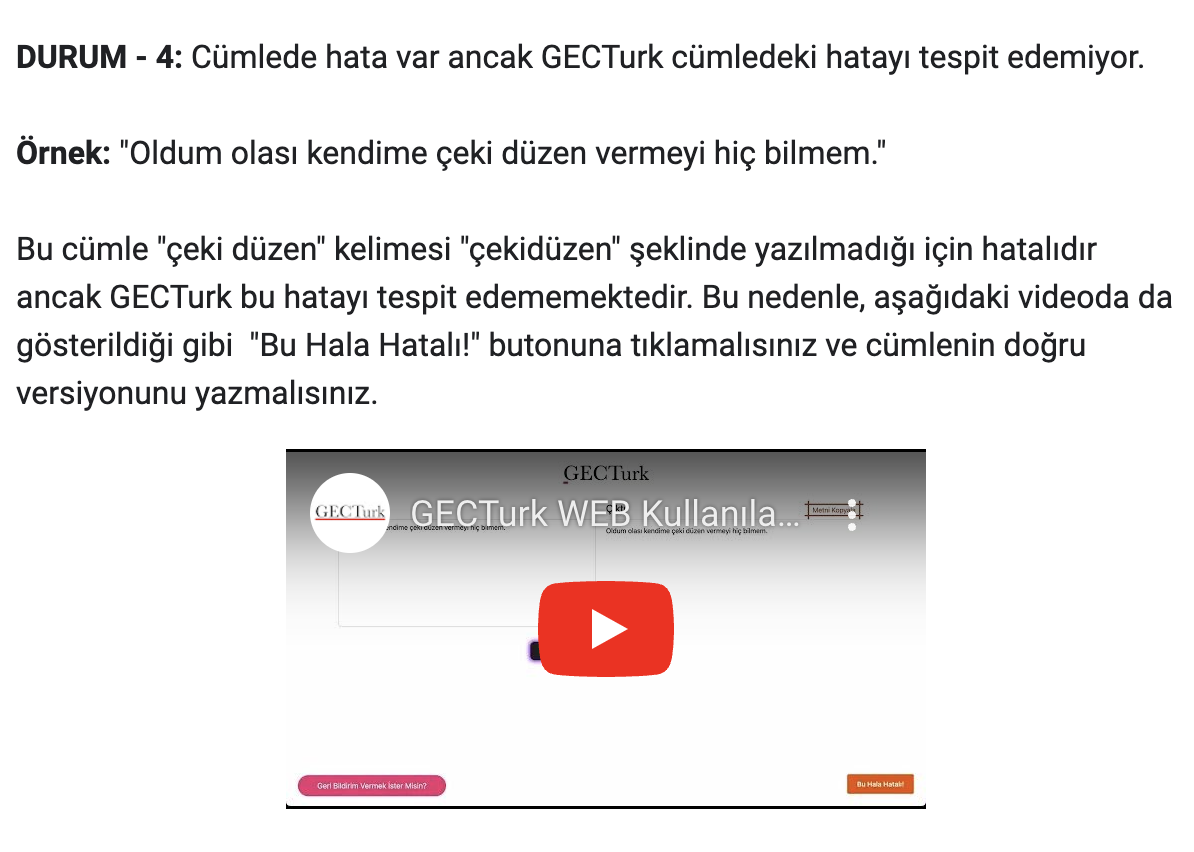}}
\caption{\textbf{CASE - 4:} There is an error in the sentence but GECTurk cannot detect it. \textbf{Example:} ``Oldum olası kendime çeki düzen vermeyi hiç bilmem.'' This sentence is incorrect because the word ``çekidüzen'' is incorrectly spelled as ``çeki düzen'', but GECTurk is unable to detect this error. Therefore, you should click on the ``This is still incorrect!'' button and type the correct version of the sentence, as shown in the video below.}
\label{fig:case-4}
\end{figure}

Participants are given 10 short sentences and are requested to input them into GECTurk WEB. To help participants understand the potential outcomes, we produced four instructional videos and showed them to the participants before they started using GECTurk WEB. Figures \ref{fig:case-1} through \ref{fig:case-4} display screenshots of each scenario along with its English transcription. Following the video demonstration, participants are directed to input each sentence and categorize it based on the possible outcomes. The full list of the 10 sentences is provided in Table \ref{table:3}.

    \begin{table*}
    \centering
    \scalebox{0.69}{
        \begin{tabular}{
          >{\centering\arraybackslash}m{2cm}
          >{\centering\arraybackslash}m{4cm}
          >{\centering\arraybackslash}m{4cm}
          >{\centering\arraybackslash}m{6cm}
          >{\centering\arraybackslash}m{2cm}
        }
        \toprule
         \textbf{Input No} & \textbf{Input} & \textbf{GECTurk WEB Output} & \textbf{Ground Truth} & \textbf{Case No}\\
        \toprule
        1 & {Dilin birey ve toplum hayatında taşıdığı önem, \textbf{anadili} öğretimini de önemli kılmaktadır.} & UNCHANGED & ... [anadili $\rightarrow$ ana dili] ... & 4 \\
        \midrule
        2 & {Onu baban görmeden hemen ortadan \textbf{kayıp et}.} & ... [kayıp et $\rightarrow$ kaybet] ... & ... [kayıp et $\rightarrow$ kaybet] ... & 1 \\
        \midrule
        3 & {Tatil yapmak \textbf{istiyrum} fakat çalışmaya devam etmem şart.} & ... [istiyrum $\rightarrow$ istiyorum] ... & ... [istiyrum $\rightarrow$ istiyorum] ... & 1 \\
        \midrule
        4 & {Bugün hep beraber gittiğimiz geziye \textbf{Ayşe'de} geldi.} & ... [Ayşe'de $\rightarrow$ Ayşe de] ... &  ... [Ayşe'de $\rightarrow$ Ayşe de] ... & 1 \\
        \midrule
        5 & \textbf{Bir takım} ansiklopediye dünyanın parasını ödedim. & [Bir takım $\rightarrow$ Birtakım] ... & UNCHANGED & 3 \\
        \midrule
        6 & {Düştüğü bu durumdan kurtulmak için \textbf{karakara} düşünüyordu.} & ... [karakara $\rightarrow$ kara kara] ... & ... [karakara $\rightarrow$ kara kara] ... & 1 \\
        \midrule
        7 & {Bugün öyle çok \textbf{yorulmuşki} hemen yattı.} & ... [yorulmuşki $\rightarrow$ yorulmuş ki] ... & ... [yorulmuşki $\rightarrow$ yorulmuş ki] ... & 1 \\
        \midrule
        8 & {Bu yaptığının elle tutulur \textbf{sebepi} yok.} & ... [sebepi $\rightarrow$ sebebi] ... & ... [sebepi $\rightarrow$ sebebi] ... & 1 \\
        \midrule
        9 & {Sanki uyurgezer biri gibi çarşıyı baştan başa adımladı.} & UNCHANGED & UNCHANGED & 2 \\
        \midrule
        10 & {\textbf{İçerde} kimsenin olmadığını gördü ve bağırmaya başladı.} & [İçerde $\rightarrow$ İçeride] ... & [İçerde $\rightarrow$ İçeride] ... & 1 \\
        \bottomrule
        \end{tabular}
    }
    \caption{The complete list of 10 sentences is given to the participants. For each sentence, participants are required to enter the \textbf{Input} and observe the \textbf{GECTurk WEB Output}. Based on this output, they decide the \textbf{Case No}. Note that participants have no access to the \textbf{Ground Truth}.}
    \label{table:3}
    \end{table*}

\FloatBarrier
\subsection{User Evaluation}
\label{ssec:second-part}
In the second part of the user study, participants are asked to complete the SUS and SUPR-Q questionnaires based on their experience in the first half of the study. Additionally, participants are asked a yes/no question regarding whether they learned or remembered a grammatical rule. The SUS questionnaire comprises ten five-level Likert scale questions, while the SUPR-Q consists of seven five-level Likert scale questions and one ten-level Likert scale question, making a total of 19 questions including the yes/no question. 

\section{Time Efficiency}
\label{sec:time-eff-app}
    \begin{figure}[!htb]
    \centering \includegraphics[width=0.5\textwidth]{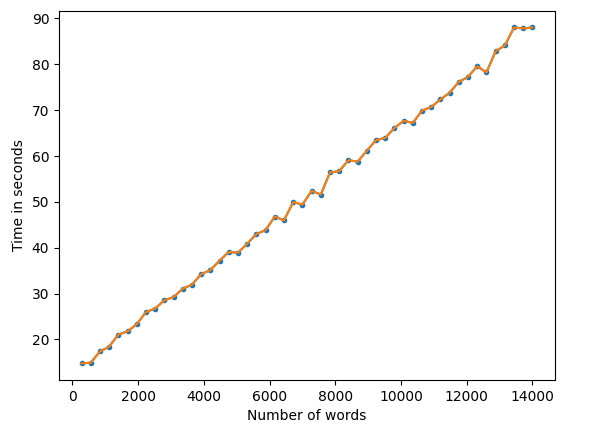} 
    \caption{The relationship between the number of words processed by the \textsc{GECTurk} model and the response time, demonstrating the model's time efficiency.}
    \label{fig:response-time}
    \end{figure}
This section highlights the model's performance in terms of time efficiency, demonstrating a linear relationship between the volume of words processed and the response time. The data suggests that the system can process up to 14,000 words in under 90 seconds, affirming its ability to scale effectively while retaining user engagement. This performance is supported by robust hardware specifications of an AWS i4i.large instance, including 2 vCPUs, 16.0 GiB of memory, and a 3.5 GHz Intel Xeon 8375C processor, which collectively ensure minimal latency even under significant text processing loads. For visual representation, see Figure~\ref{fig:response-time}.

\end{document}